\documentclass{article}
\usepackage{spconf,amsmath,graphicx}
\usepackage{graphicx}

\usepackage{subfigure}
\usepackage{amsmath}
\usepackage{booktabs}
\usepackage{amsfonts}
\usepackage{color}
\usepackage{bm}
\usepackage{arydshln}
\usepackage{multirow}
\usepackage{enumitem}
\usepackage{extarrows}
\usepackage{amssymb}
\usepackage{caption} % DO NOT CHANGE THIS AND DO NOT ADD ANY OPTIONS TO IT
\usepackage{newfloat}
\usepackage{hyperref}

\title{MoDeRNN: Towards Fine-grained Motion Details for spatiotemporal predictive learning}

\name{Zenghao Chai$^{1}$ \qquad Zhengzhuo Xu$^{1}$ \qquad Chun Yuan$^{1,2}$\sthanks{Corresponding Author}}
\address{$^1$ Shenzhen International Graduate School, Tsinghua University, Shenzhen, China  \\
      $^2$ Peng Cheng Laboratory, Shenzhen, China \\
zenghaochai@gmail.com; xzz20@mails.tsinghua.edu.cn; yuanc@sz.tsinghua.edu.cn}

\begin{document}
%\ninept
%
\maketitle
\begin{abstract}

Spatiotemporal predictive learning (ST-PL) aims at predicting the subsequent frames via limited observed sequences, and it has broad applications in the real world. However, learning representative spatiotemporal features for prediction is challenging. Moreover, chaotic uncertainty among consecutive frames exacerbates the difficulty in long-term prediction. This paper concentrates on improving prediction quality by enhancing the correspondence between the previous context and the current state. We carefully design Detail Context Block (DCB) to extract fine-grained details and improve the isolated correlation between upper context state and current input state. We integrate DCB with standard ConvLSTM and introduce Motion Details RNN (MoDeRNN) to capture fine-grained spatiotemporal features and improve the expression of latent states of RNNs to achieve significant quality. Experiments on Moving MNIST and Typhoon datasets demonstrate the effectiveness of the proposed method. MoDeRNN outperforms existing state-of-the-art techniques qualitatively and quantitatively with lower computation loads.

\end{abstract}

\section{Introduction}

\textbf{S}patio-\textbf{T}emporal \textbf{P}redictive \textbf{L}earning (ST-PL) is challenging with broad applications in predictive learning, e.g., physical object movement \cite{lerer2016learning,DBLP:conf/nips/FinnGL16,DBLP:conf/nips/SuBKHKA20,DBLP:conf/eccv/BhagatUYL20}, meteorological prediction \cite{shi2015convolutional,DBLP:journals/corr/abs-2103-02243,wang2019memory,DBLP:conf/kdd/GengLLJXZYLZ19}. It aims to predict future sequences based on limited observed frames. The difficulty of ST-PL lies in the chaotic motion trends and profound dynamic changes. Hence it's necessary and crucial to build a proper corresponding between current input frames and previous observations, and integrate the motion trends for subsequent prediction.

Recent years have achieved impressive progress in ST-PL, plenty of novel approaches  \cite{shi2015convolutional,DBLP:conf/nips/ShiGL0YWW17,wang2017predrnn,DBLP:journals/corr/abs-2103-09504,wang2018predrnn++,wang2019memory,DBLP:conf/iclr/WangJYLLF19,lin2020self} are proposed for long-term prediction. As one of the most popular branches, RNN \cite{hochreiter1997long} or LSTM \cite{werbos1990backpropagation} plays an important role as a mainstream model. These methods have shown impressive results in ST-PL and made persistent progress. RNN and LSTM based models predict subsequent frames in an auto-regressive mode, i.e., stacked RNN layers embed input features obtained by CNN layers into latent states and update hidden states by the elaborate designed process to obtain output states, and decode to obtain the next timestamp frames.

However, when rethinking the calculation process of ConvLSTM \cite{shi2015convolutional,DBLP:conf/nips/ShiGL0YWW17} and its extensions \cite{wang2017predrnn,wang2018predrnn++,wang2019memory,lin2020self}, it's intuitive that the input state and upper context state show isolated correspondence in the process of RNN layers. The two states in previous models are only correlated by CNN layers and channel-wise addition operation. Hence models confront the two severe dilemmas that will lead to worse prediction results in long-term prediction: 1. The increasing models' depth and complexity exacerbate the declination of correlations between the current input and upper context, making it even difficult to build correct correspondence between the current frame and upper context. 2. CNNs can hardly capture fine-grained features that contain abundant details for prediction, limiting the ability to consider detailed features of latent states.

On top of the aforementioned, the current frame states are highly correlated to its neighbors of specific regions, i.e., the next timestamp frames in a region are related to both itself and its neighbor subject movements. The fine-grained local information is crucial for long-term prediction. To improve the correlation and the detailed local information between input and context, we propose Motion Details RNN (MoDeRNN) to tackle the above challenges in ST-PL effectively.

MoDeRNN contains the carefully designed Detail Context Block (DCB), which weights input and context states to highlight the spatiotemporal details for subsequent prediction. In specific, to obtain latent spatiotemporal trends among different neighbors, DCB utilizes various perceptual fields CNN layers to capture regions corresponding to input states and context states, and updates the corresponding context state and input state iteratively with rich correlations. As a result, the proposed MoDeRNN enables to capture fine-grained locals to persist correlations among RNN layers and achieves remarkable satisfactory prediction performance. 

\begin{figure*}[ht!]
  \centering
  \includegraphics[width=1\linewidth ]{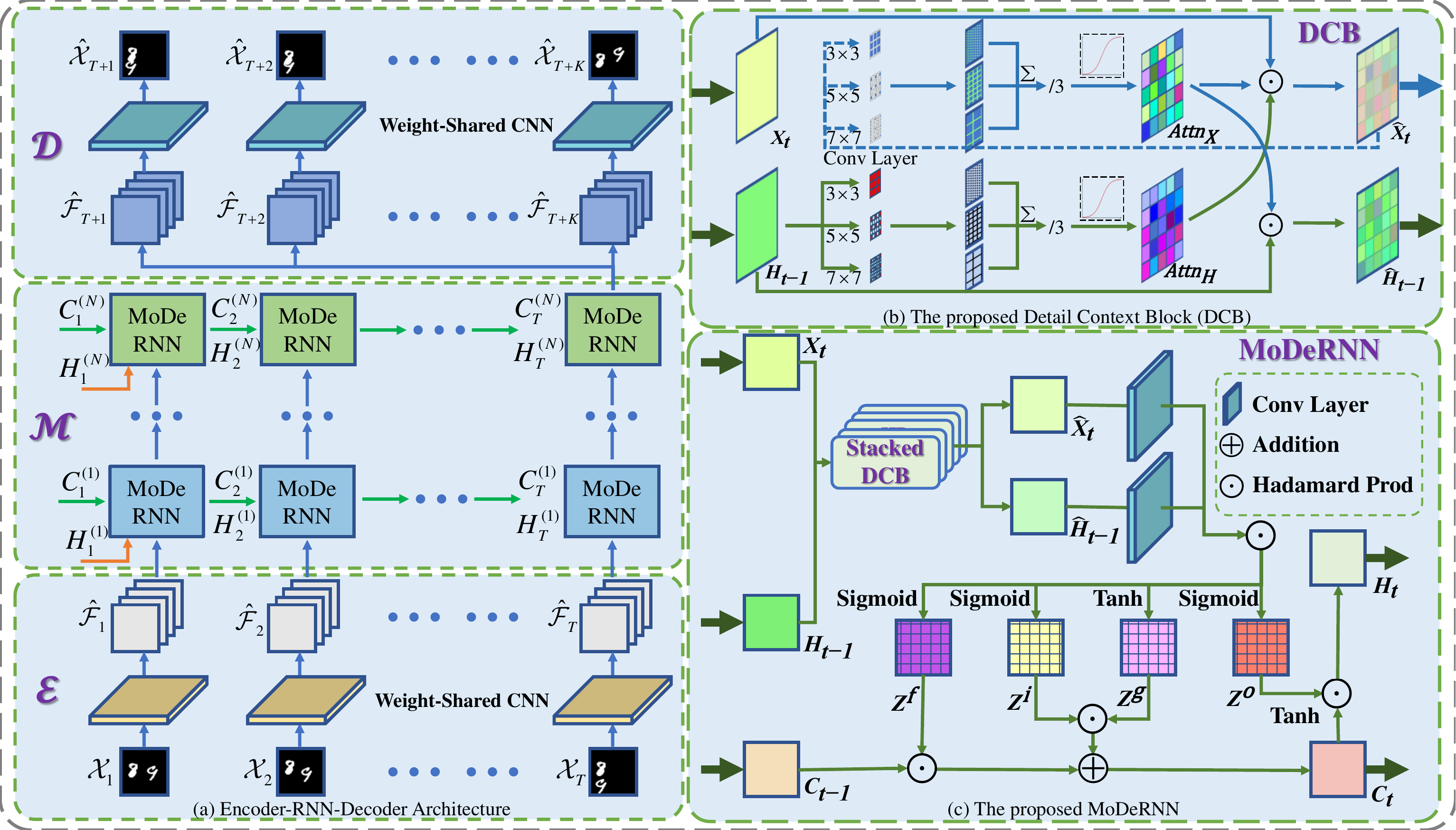}
  \caption{Overview of the proposed Method. (a): mainstream architecture for ST-PL; (b): pipeline of DCB; (c): pipeline of the proposed MoDeRNN.}
  \label{fig::MoDeRNN}
  \vspace{-0.5cm}
\end{figure*}

In summary, our main contributions are two-fold:
\begin{itemize}[leftmargin=*]% itemsep=2pt,topsep=0pt,parsep=0pt,
    \item We construct Detail Context Block to capture fine-grained local details and update context states with dense correlations. We analyze the essential of context attention over fine-grained regions for prediction, and propose MoDeRNN towards fined-grained detailed prediction quality.
    \item We validate that the proposed MoDeRNN well captures trend regions of given frames to obtain better context and input correlations, and achieves distinguish performance gains compared to previous methods with fewer params on two representative datasets.
\end{itemize}

% \vspace{-3em}
\section{Methodology}
\subsection{Model Architecture}

The RNN-based models are universally used approaches for ST-PL, with common Encoder-RNN-Decoder architecture \cite{shi2015convolutional,wang2017predrnn,wang2018predrnn++,wang2019memory,lin2020self} as Fig.\ref{fig::MoDeRNN}(a) shows. The given frames are encoded by 2D CNN \cite{lecun1995convolutional} encoder $\mathcal{E}$ in step-by-step mode, then the obtained features $[\hat{\mathcal{F}}_{1:T}]$ serve as the input of $N$-layer LSTMs denoted as $\mathcal{M}$ to generate high-order spatiotemporal features of given sequences and output states $[\hat{\mathcal{F}}_{T+1:T+K}]$. Ultimately, the output states are  decoded by 2D CNN decoder $\mathcal{D}$ iteratively and thus generate the next $K$ frames $[\hat{\mathcal{X}}_{T+1:T+K}]$. The mathematics pipeline is illustrated as Eq.\ref{equ:pipeline}.

\begin{equation}
  \begin{split}
    &[\hat{\mathcal{X}}_{1:T}]=\mathcal{E}([\mathcal{X}_{1:T}])\\
    &[\hat{\mathcal{F}}_{T+1:T+K}]=\mathcal{M}([\hat{\mathcal{X}}_{1:T}])\\
    &[\hat{\mathcal{X}}_{T+1:T+K}]=\mathcal{D}([\hat{\mathcal{F}}_{T+1:T+K}]) \\
  \end{split}
  \label{equ:pipeline}
\end{equation}
In this paper, we keep the encoder $\mathcal{E}$ and decoder $\mathcal{D}$ consistent with previous work mentioned above(\cite{wang2017predrnn,lin2020self}, etc.). Namely, they are both $1\times 1$ kernel CNN layers. The crucial target is to make representative high-order spatiotemporal features on RNNs, while there are issues worthy of consideration.

\subsection{The proposed DCB}
The abundant context feature is hard to obtain due to the limited operation between the current input and the previous context state in RNNs. In na\"ive ConvLSTM, the correlation of input state and context state is operated by CNN layers and add operation, while the subsequent state updates don't involve the interaction of the two states. Therefore, their relationship remains independent in the following operation, which easily leads to the loss of information in the prediction results. Intuitively, the output frames will get increasingly worse prediction quality, especially in detail parts. 

Considering the importance of improving the correlation between neighbors and context states, we propose Detail Context Block (DCB) to extract fine-grained local features of current input state $X_t$ and context state $H_{t-1}$, and utilize the proposed context interaction approach to improve the correlations between the upper context state and current input state. The detailed architecture of DCB is illustrated in Fig.\ref{fig::MoDeRNN}(b). 

In specific, to extract fine-grained local features, we utilize CNN layers with different perceptual fields to comprehensively focus on detailed motion regions of context state and input state, respectively, and use iterative weight correlate operations to improve the isolated correlation between the two states. DCB consists of the following steps:

\noindent \textbf{Step 1.} To obtain specific regions in current input state $X_t$ that are essential for prediction, we comprehensively consider the influence of locals by generating an attention weight map $Attn_H$ of upper context via multi-kernel CNN layers that capture the context features, then obtain the mean local features that indicates the potential movement trend in the following timestamp. We adopt $Sigmoid$ function $\sigma$ to normalize the weight map into $(0,1)$, and reweight the input feature $X_t$ by the Hadamard product to highlight the important part of the input state. Finally, we multiply the weight map by a constant scale 
factor $s$ to avoid getting increasingly smaller.

\noindent \textbf{Step 2.} We encourage upper context $H_{t-1}$ updates by considering the trends of current input, i.e., enforce $H_{t-1}$ enlighten the fine-grained motion details and weaken the negligible parts with lower expression simultaneously. We update $H_{t-1}$ by multiplying an input-related attention weight map $Attn_X$ to extract motion concentration for prediction by Hadamard product. The weight map is calculated the same as Step 1, i.e., capture the detailed context motion features by multi-kernel size CNN layers and activation function $\sigma$ with scale factor $s$. Then, the updated context state $\hat{H}_{t-1}$ and input state $\hat{X}_t$ are obtained with rich spatiotemporal features.

\subsection{Overview of MoDeRNN}

Considering improving the expression ability in detail regions for ST-PL, we integrate DCB with na\"ive ConvLSTM to compose the proposed MoDeRNN as Fig.\ref{fig::MoDeRNN}(c), a new spatiotemporal prediction model towards fine-grained details. Formally, MoDeRNN can be expressed as follows:

Firstly, we utilize DCB to capture fine-grained detail spatiotemporal features and update current input state $X_t$ and upper context state $H_{t-1}$. Then, to further improve the correlations between the two states, we use $m$ stacked DCB with kernel size varies from $k\in\{3,5,7\}$ to improve the expression ability of MoDeRNN with more details.
\begin{equation}
    \begin{split}
    % update H_t-1 and X_t
    &Attn_H=\sigma \left( \sum_{i}^{k} W_{h}^{i\times i}\star H_{t-1} / |k|\right)\\
    &\hat{X}_t = s \times Attn_H \times H_{t-1} \\
    &Attn_X=\sigma \left( \sum_{i}^{k} W_{x}^{i\times i}\star \hat{X}_{t-1}/|k| \right) \\
    &\hat{H}_{t-1} = s \times Attn_X \times \hat{X}_{t-1} \\
    \end{split}
    \label{equ:DCB}
\end{equation}
where $W_{h}^{i \times i}$ and $W_{x}^{i\times i}$ represent $i\times i$ kernel CNN layers for $H_{t-1}$ and $\hat{X}_t$, respectively. $s$ represents the scale factor and $\sigma$ indicates $Sigmoid$ activation function.

Secondly, we utilize the updated $\hat{X}_t$ and $\hat{H}_{t-1}$ to obtain the detailed output state $H_{t}$ and memory state $C_t$. In the last layer of MoDeRNN, the final output state $H_{t}$ is decoded to generate the final output frame of the next timestamp.
\begin{equation}
    \begin{split}
    % update H_t-1 and X_t
    &g_t=\tanh (W_{xg} \star \hat{X}_t+W_{hg} \star \hat{H}_{t-1}+b_g)\\
    &i_t=\sigma (W_{xi} \star \hat{X_t} + W_{hi} \star \hat{H}_{t-1}+b_i)\\
    &f_t=\sigma(W_{xf} \star \hat{X_t}+W_{hf} \star \hat{H}_{t-1}+b_f) \\
    &C_t=f_t\circ C_{t-1}+i_t\circ g_t\\
    &o_t=\sigma(W_{xo} \star \hat{X}_t +W_{ho}\star \hat{H}_{t-1}+b_o)\\
    &H_t=o_t\circ \tanh (C_t) \\
    \end{split}
    \label{equ:MoDeRNN}
\end{equation}
where $W_{xg},W_{hg},W_{xi},W_{hi},W_{xf},W_{hf}$ are $5\times 5$ kernel CNN layers for gate operation.
\section{Experiment}

\vspace{-5pt}
\subsection{Experiment Details}
We implement the proposed model by Pytorch \cite{paszke2019pytorch}, train and test it on a single RTX 2080Ti. For fair comparisons, we use $4$-layer MoDeRNN units with $64$-dim hidden states consistent with previous work. We set the mini-batch as $32$ with the initial learning rate as $0.001$. We also adopt scheduled sampling \cite{bengio2015scheduled} and layer normalization \cite{ba2016layer} for better results. During training, we use $L_1+L_2$ loss with AdamW \cite{loshchilov2018fixing} optimizer. Code is avaliable at \href{https://github.com/czh-98/MoDeRNN}{https://github.com/czh-98/MoDeRNN}.

\vspace{-5pt}
\subsection{Dataset}

\noindent \textbf{Moving MNIST.} Moving MNIST \cite{srivastava2015unsupervised} is a widespread benchmark for depicting $2$ digits’ movement with constant velocity. It contains $64\times 64\times 1$ consecutive frames with $10$ for input and $10$ for prediction, $10,000$ randomly generated sequences for training and $10,000$ fixed parts for testing.

\noindent \textbf{Typhoon.} Typhoon dataset is a meteorology radar data released by CEReS \cite{DBLP:journals/remotesensing/YamamotoIHT20}. We resize the images into $64 \times 64 \times 1$ resolution and normalize to $[0,1]$, then split the generated sequences into train and test sets. We use the given $8$-hour observation data to predict the next $4$ hours, with $1,809$ sequences for training and $603$ sequences for testing.

\vspace{-5pt}
\subsection{Comparisons on Moving MNIST}

\begin{figure}[t!]
  \centering
  \includegraphics[width=1\linewidth]{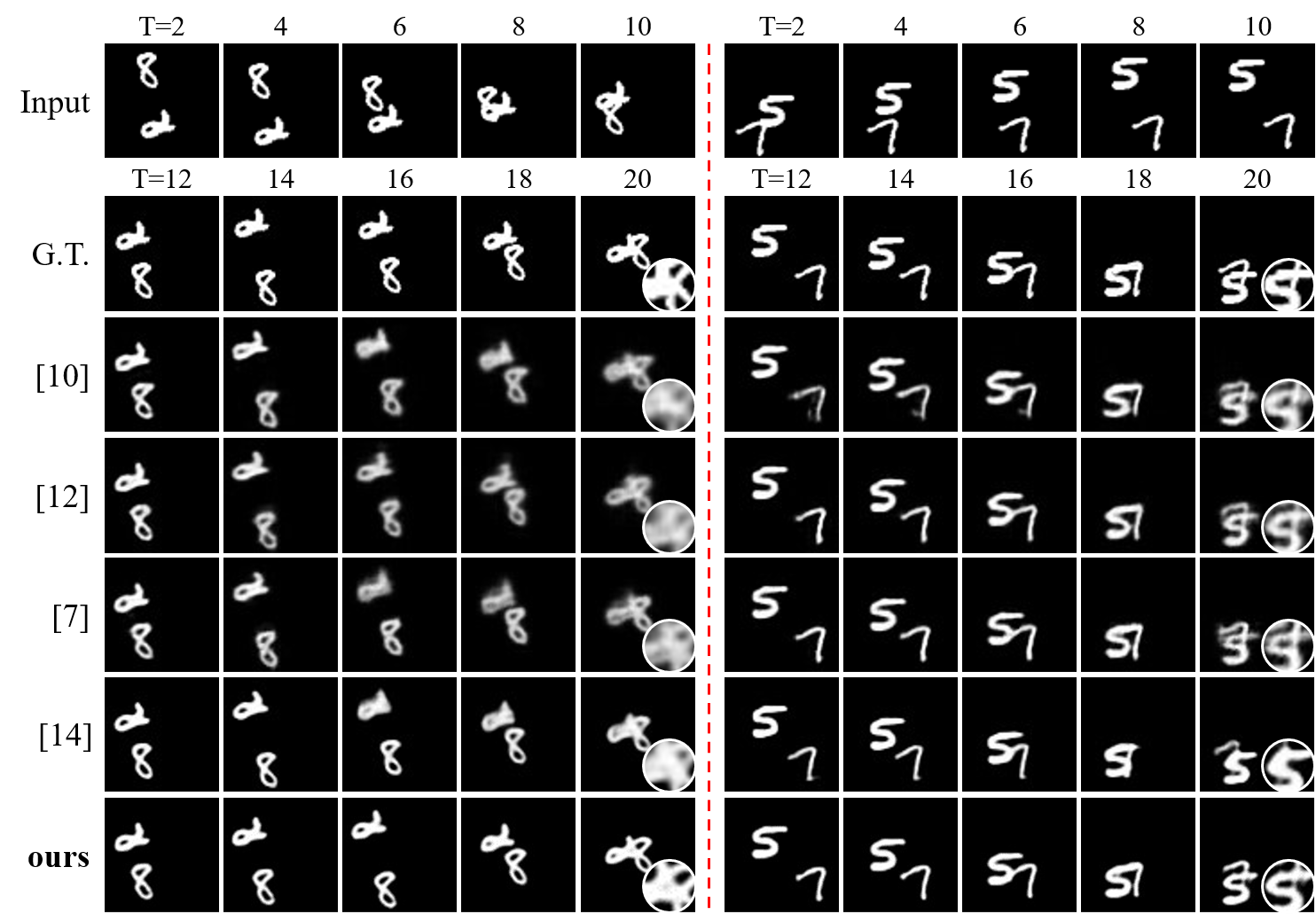}
  \vspace{-20pt}
  \caption{Qualitative comparisons of previous SOTA models on Moving MNIST test set at $80,000$ iterations.}
  \label{fig::mmnist}
  \vspace{-10pt}
\end{figure}

We set $80,000$ iterations consistent with previous work (\cite{wang2017predrnn} etc.).  We use PSNR, SSIM, MSE, and MAE for quantitative comparisons. The higher SSIM / PSNR and lower MSE / MAE indicate better performance. Results in Tab.\ref{tab::mmnist} demonstrate the superiority of our method on Moving MNIST dataset in all above metrics, improving $\bm{9.62\%}$ and $\bm{2.52\%}$ on PSNR and SSIM, and reducing $\bm{12.03\%}$ and $\bm{27.46\%}$ on MSE and MAE respectively compared with SA-ConvLSTM \cite{lin2020self}, while achieving lower computational loads. Fig.\ref{fig::mmnist} shows that MoDeRNN well preserves the variation details over digits, especially deals with the trajectory of overlaps and maintains the clarity over time. In contrast, other methods confront severe blurry challenges and are incapable of dealing with overlapped digits.
% \vspace{-0.5cm}

% \vspace{-10pt}
\begin{table}[t!]
\resizebox{\linewidth}{!}{%
\begin{tabular}{l|c|cccc}
\toprule \hline
\textbf{Models} & \textbf{\# Params} & \textbf{PSNR $\uparrow$}   & \textbf{SSIM $\uparrow$}  & \textbf{MSE $\downarrow$}  & \textbf{MAE $\downarrow$}  \\ \midrule 
DDPAE \cite{DBLP:conf/nips/HsiehLHLN18}        & -                  & 21.170          & 0.922          & 38.9          & 90.7          \\
CrevNet \cite{DBLP:conf/iclr/YuLEF20}        & -                  & -               & 0.928          & 38.5          & -             \\
PDE-Driven \cite{2021pdedriven}     & -                  & 21.760          & 0.909          & -             & -             \\
PredRNN \cite{wang2017predrnn}        & 13.799 M           & 19.603          & 0.867          & 56.8          & 126.1         \\
PredRNN++ \cite{wang2018predrnn++}      & 13.237 M           & 20.239          & 0.898          & 46.5          & 106.8         \\
MIM* \cite{wang2019memory}           & 27.971 M           & 20.678          & 0.910          & 44.2          & 101.1         \\
E3D-LSTM \cite{DBLP:conf/iclr/WangJYLLF19}       & 38.696 M           & 20.590          & 0.910          & 41.7          & 87.2          \\
SA-ConvLSTM \cite{lin2020self}    & 10.471 M           & 20.500          & 0.913          & 43.9          & 94.7          \\ \midrule
\textbf{MoDeRNN (ours)}  & \textbf{4.590 M}   & \textbf{22.472} & \textbf{0.936} & \textbf{30.6} & \textbf{68.7} \\ \hline \bottomrule
\end{tabular}
}
\vspace{-5pt}
\caption{Quantitative comparisons of previous SOTA models on Moving MNIST test set. All models predict $10$ frames by observing $10$ previous frames.}
\label{tab::mmnist}
\vspace{-20pt}
% \vspace{-0.8cm}
\end{table}

\begin{figure}[b!]
\vspace{-15pt}
% \vspace{-0.1cm}
  \centering
  \includegraphics[width=1\linewidth]{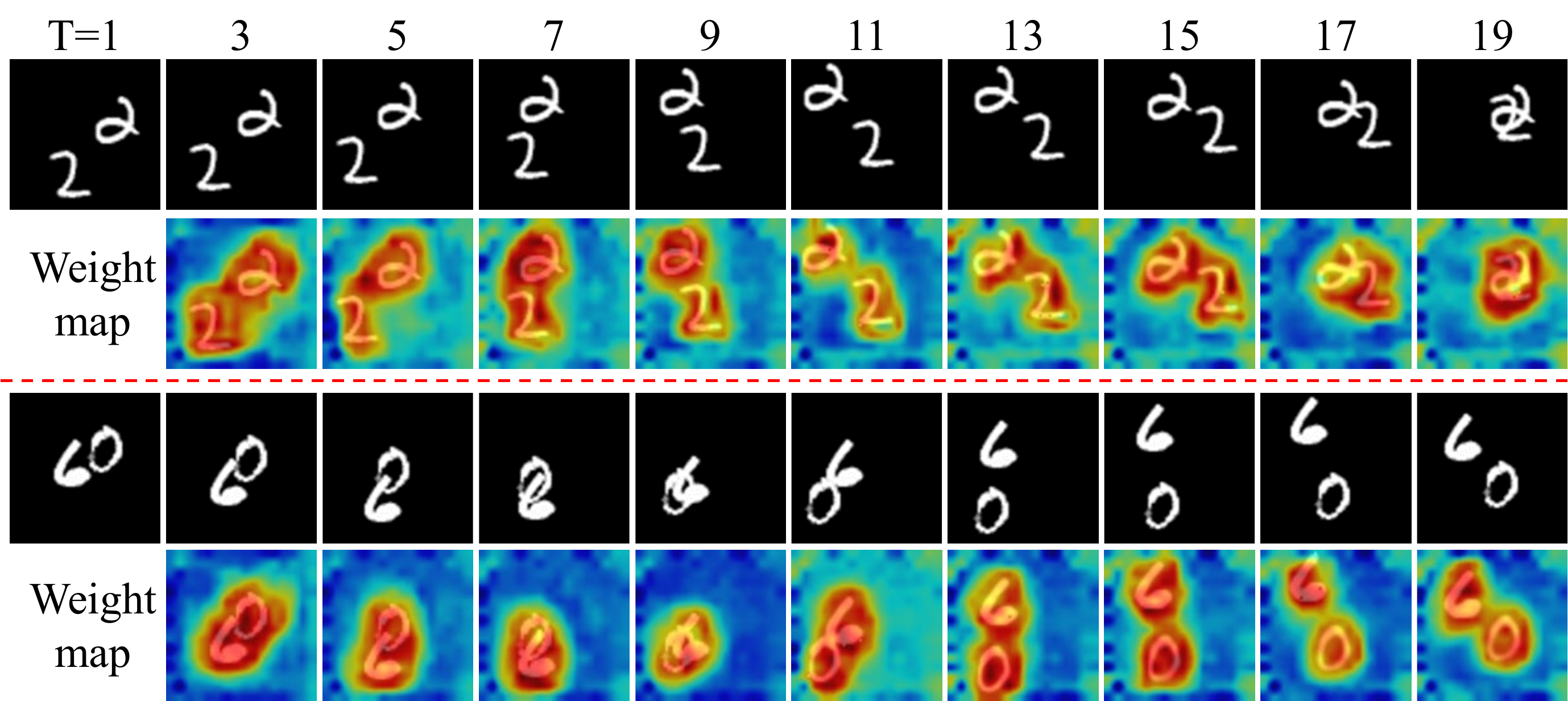}
    \vspace{-18pt}
  \caption{Visualization of MoDeRNN on Moving MNIST test set of the last layer, the warm colors
indicate higher weights.}
  \label{fig::attn}
%   \vspace{-10pt}
\end{figure}

Fig.\ref{fig::attn} illustrates the weight map over consecutive timestamps. MoDeRNN focuses on fine-grained local details for subsequent prediction and can even handle overlap scenarios.

\subsection{Comparisons on Typhoon}

\begin{figure}[t!]
  \centering
  \includegraphics[width=1\linewidth]{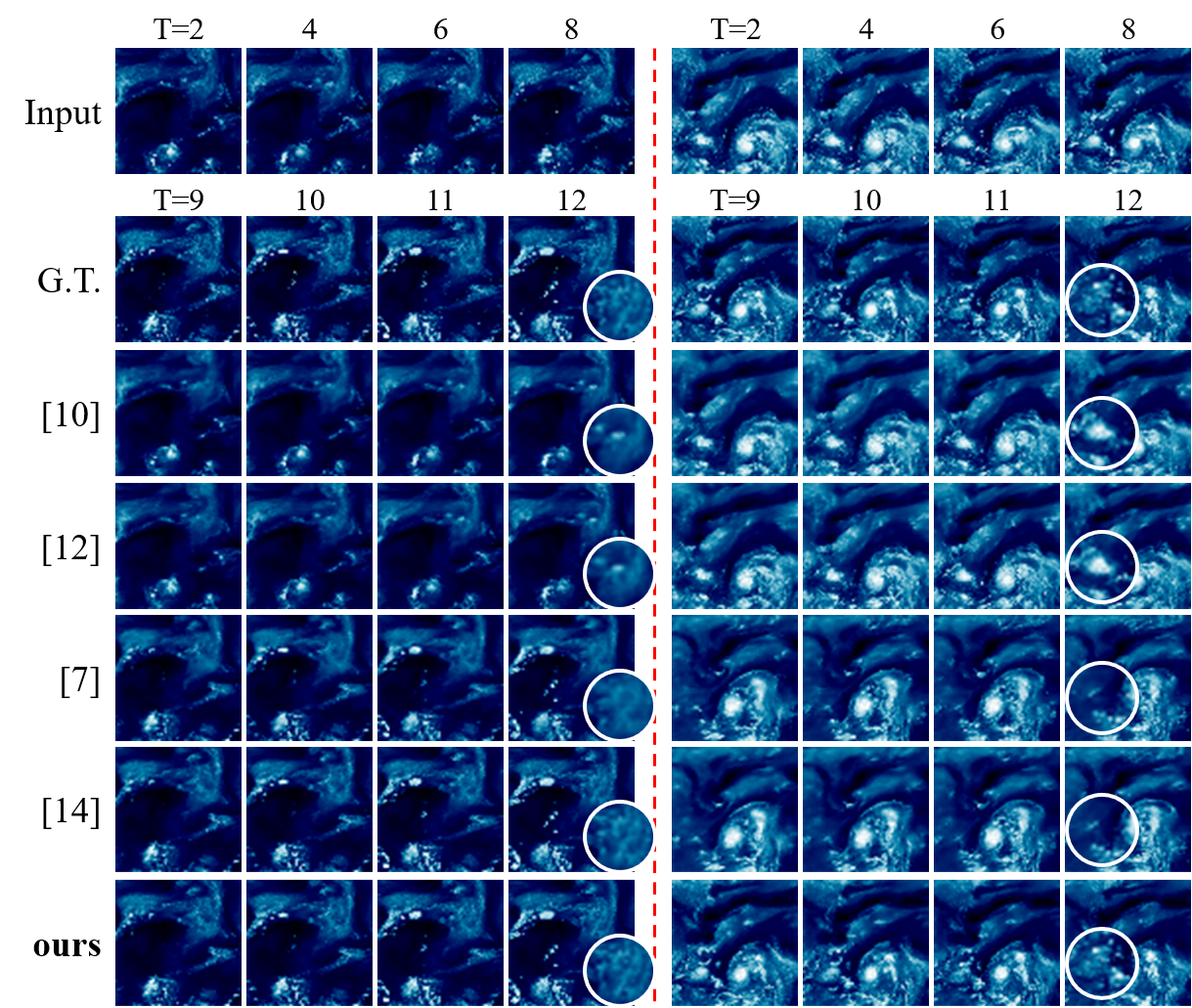}
    \vspace{-15pt}
  \caption{Qualitative comparisons of previous SOTA models on Typhoon test set at $100,000$ iterations.}
  \label{fig::typhoon}
%   \vspace{-0.3cm}
\end{figure}

We train the proposed models for $100,000$ iterations and make fair comparisons with previous methods \cite{shi2015convolutional,wang2017predrnn,wang2018predrnn++,wang2019memory,lin2020self}. Frame-wise PSNR, SSIM, MSE, and MAE are adopted to evaluate these models’ performance qualitatively and quantitatively, corresponding to Fig.\ref{fig::typhoon} and Tab.\ref{tab::exptyphoon}.

Tab.\ref{tab::exptyphoon} and Fig.\ref{fig::typhoon} demonstrate the proposed method outperforms existing techniques quantitatively and qualitatively. MoDeRNN is the only model that performs well in the detail texture of over timestamps, which enables to preserve and predict the potential trend of meteorological information.

\begin{table}[t!]
\setlength{\tabcolsep}{10pt}
\centering
\resizebox{\linewidth}{!}{%
\begin{tabular}{l|l|l|l|l|l|l|l|l|l|l|l|l|l|l|l|l}
\toprule
\hline
\textbf{Models}                                                & \multicolumn{4}{c|}{\textbf{PSNR} $\uparrow$}                  & \multicolumn{4}{c|}{\textbf{SSIM} $\uparrow$}                 & \multicolumn{4}{c|}{\textbf{MSE} $\downarrow$}               & \multicolumn{4}{c}{\textbf{MAE} $\downarrow$}               \\ \hline

ConvLSTM \cite{shi2015convolutional} & \multicolumn{4}{l|}{26.353}                           & \multicolumn{4}{l|}{0.851}                           & \multicolumn{4}{l|}{10.43}                          & \multicolumn{4}{l}{119.6}                          \\
PredRNN \cite{wang2017predrnn}      & \multicolumn{4}{l|}{27.637}                           & \multicolumn{4}{l|}{0.887}                           & \multicolumn{4}{l|}{7.71}                           & \multicolumn{4}{l}{107.3}                          \\
PredRNN++ \cite{wang2018predrnn++}   & \multicolumn{4}{l|}{28.287}                           & \multicolumn{4}{l|}{0.891}                           & \multicolumn{4}{l|}{6.72}                           & \multicolumn{4}{l}{114.5}                          \\
MIM* \cite{wang2019memory}        & \multicolumn{4}{l|}{26.721}                           & \multicolumn{4}{l|}{0.893}                           & \multicolumn{4}{l|}{9.14}                           & \multicolumn{4}{l}{132.2}                          \\
SA-ConvLSTM \cite{lin2020self}    & \multicolumn{4}{l|}{28.456}                           & \multicolumn{4}{l|}{0.898}                           & \multicolumn{4}{l|}{7.07}                           & \multicolumn{4}{l}{94.2}                           \\ \midrule
\textbf{MoDeRNN (ours)}                  & \multicolumn{4}{l|}{\textbf{29.446}} & \multicolumn{4}{l|}{\textbf{0.910}} & \multicolumn{4}{l|}{\textbf{6.06}} & \multicolumn{4}{l}{\textbf{83.1}} \\ \hline
  \bottomrule
\end{tabular}}
  \vspace{-5pt}
\caption{ Quantitative comparisons of previous SOTA models on Typhoon test set. All models predict the next $4$ frames via $8$ observed meteorological data.}
\label{tab::exptyphoon}
\vspace{-0.5cm}
\end{table}
\section{Conclusion}
This paper introduces the novel MoDeRNN for ST-PL, which focuses on tackling the challenging motion trends towards detailed prediction. We propose MoDeRNN to capture fine-grained spatiotemporal latent features to improve the prediction quality in long-term prediction. 

In detail, we propose DCB to make latent states well interacted with fine-grained motion details and ensure the prediction results keep consistent clarity. We demonstrate that MoDeRNN achieves satisfactory performance compared to mainstream methods with the lower computational load on $2$ representative datasets.

% \newpage
\bibliographystyle{IEEE}
\bibliography{IEEE}

\begin{thebibliography}{10}

\bibitem{lerer2016learning}
Adam Lerer, Sam Gross, and Rob Fergus,
\newblock ``Learning physical intuition of block towers by example,''
\newblock pp. 430--438, 2016.

\bibitem{DBLP:conf/nips/FinnGL16}
Chelsea Finn, Ian~J. Goodfellow, and Sergey Levine,
\newblock ``Unsupervised learning for physical interaction through video
  prediction,''
\newblock in {\em NeurIPS 2016}, 2016, pp. 64--72.

\bibitem{DBLP:conf/nips/SuBKHKA20}
Jiahao Su, Wonmin Byeon, Jean Kossaifi, et~al.,
\newblock ``Convolutional tensor-train {LSTM} for spatio-temporal learning,''
\newblock in {\em NeurIPS 2020}, 2020.

\bibitem{DBLP:conf/eccv/BhagatUYL20}
Sarthak Bhagat, Shagun Uppal, Zhuyun Yin, et~al.,
\newblock ``Disentangling multiple features in video sequences using gaussian
  processes in variational autoencoders,''
\newblock in {\em ECCV 2020}. 2020, pp. 102--117, Springer.

\bibitem{shi2015convolutional}
Xingjian Shi, Zhourong Chen, Hao Wang, et~al.,
\newblock ``Convolutional lstm network: A machine learning approach for
  precipitation nowcasting,''
\newblock {\em Advances in neural information processing systems}, pp.
  802--810, 2015.

\bibitem{DBLP:journals/corr/abs-2103-02243}
Haixu Wu, Zhiyu Yao, Mingsheng Long, et~al.,
\newblock ``Motionrnn: {A} flexible model for video prediction with
  spacetime-varying motions,''
\newblock {\em CoRR}, vol. abs/2103.02243, 2021.

\bibitem{wang2019memory}
Yunbo Wang, Jianjin Zhang, Hongyu Zhu, et~al.,
\newblock ``Memory in memory: {A} predictive neural network for learning
  higher-order non-stationarity from spatiotemporal dynamics,''
\newblock in {\em CVPR 2019}, 2019, pp. 9154--9162.

\bibitem{DBLP:conf/kdd/GengLLJXZYLZ19}
Yangli{-}ao Geng, Qingyong Li, Tianyang Lin, et~al.,
\newblock ``Lightnet: {A} dual spatiotemporal encoder network model for
  lightning prediction,''
\newblock in {\em SIGKDD 2019}. 2019, pp. 2439--2447, {ACM}.

\bibitem{DBLP:conf/nips/ShiGL0YWW17}
Xingjian Shi, Zhihan Gao, Leonard Lausen, et~al.,
\newblock ``Deep learning for precipitation nowcasting: {A} benchmark and {A}
  new model,''
\newblock in {\em NeurIPS 2017}, 2017, pp. 5617--5627.

\bibitem{wang2017predrnn}
Yunbo Wang, Mingsheng Long, Jianmin Wang, et~al.,
\newblock ``Predrnn: Recurrent neural networks for predictive learning using
  spatiotemporal lstms,''
\newblock in {\em NeurIPS 2017}, 2017, pp. 879--888.

\bibitem{DBLP:journals/corr/abs-2103-09504}
Yunbo Wang, Haixu Wu, Jianjin Zhang, et~al.,
\newblock ``Predrnn: {A} recurrent neural network for spatiotemporal predictive
  learning,''
\newblock {\em CoRR}, vol. abs/2103.09504, 2021.

\bibitem{wang2018predrnn++}
Yunbo Wang, Zhifeng Gao, Mingsheng Long, et~al.,
\newblock ``Predrnn++: Towards {A} resolution of the deep-in-time dilemma in
  spatiotemporal predictive learning,''
\newblock pp. 5110--5119, 2018.

\bibitem{DBLP:conf/iclr/WangJYLLF19}
Yunbo Wang, Lu~Jiang, Ming{-}Hsuan Yang, et~al.,
\newblock ``Eidetic 3d {LSTM:} {A} model for video prediction and beyond,''
\newblock in {\em ICLR 2019}, 2019.

\bibitem{lin2020self}
Zhihui Lin, Maomao Li, Zhuobin Zheng, et~al.,
\newblock ``Self-attention convlstm for spatiotemporal prediction,''
\newblock in {\em AAAI 2020}, 2020, pp. 11531--11538.

\bibitem{hochreiter1997long}
Sepp Hochreiter and J{\"u}rgen Schmidhuber,
\newblock ``Long short-term memory,''
\newblock {\em Neural computation}, , no. 8, pp. 1735--1780, 1997.

\bibitem{werbos1990backpropagation}
Paul~J Werbos,
\newblock ``Backpropagation through time: what it does and how to do it,''
\newblock {\em Proceedings of the IEEE}, , no. 10, pp. 1550--1560, 1990.

\bibitem{lecun1995convolutional}
Yann LeCun, Yoshua Bengio, et~al.,
\newblock ``Convolutional networks for images, speech, and time series,''
\newblock {\em The handbook of brain theory and neural networks}, vol. 3361,
  no. 10, pp. 1995, 1995.

\bibitem{paszke2019pytorch}
Adam Paszke, Sam Gross, Francisco Massa, et~al.,
\newblock ``Pytorch: An imperative style, high-performance deep learning
  library,''
\newblock in {\em NeurIPS 2019}, 2019, pp. 8024--8035.

\bibitem{bengio2015scheduled}
Samy Bengio, Oriol Vinyals, Navdeep Jaitly, et~al.,
\newblock ``Scheduled sampling for sequence prediction with recurrent neural
  networks,''
\newblock pp. 1171--1179, 2015.

\bibitem{ba2016layer}
Jimmy~Lei Ba, Jamie~Ryan Kiros, and Geoffrey~E Hinton,
\newblock ``Layer normalization,''
\newblock {\em arXiv preprint arXiv:1607.06450}, 2016.

\bibitem{loshchilov2018fixing}
Ilya Loshchilov and Frank Hutter,
\newblock ``Fixing weight decay regularization in adam,''
\newblock {\em CoRR}, 2017.

\bibitem{srivastava2015unsupervised}
Nitish Srivastava, Elman Mansimov, and Ruslan Salakhutdinov,
\newblock ``Unsupervised learning of video representations using lstms,''
\newblock in {\em ICML 2015}, 2015, pp. 843--852.

\bibitem{DBLP:journals/remotesensing/YamamotoIHT20}
Yuhei Yamamoto, Kazuhito Ichii, Atsushi Higuchi, et~al.,
\newblock ``Geolocation accuracy assessment of himawari-8/ahi imagery for
  application to terrestrial monitoring,''
\newblock {\em Remote. Sens.}, vol. 12, no. 9, pp. 1372, 2020.

\bibitem{DBLP:conf/nips/HsiehLHLN18}
Jun{-}Ting Hsieh, Bingbin Liu, De{-}An Huang, et~al.,
\newblock ``Learning to decompose and disentangle representations for video
  prediction,''
\newblock in {\em NeurIPS 2018}, 2018, pp. 515--524.

\bibitem{DBLP:conf/iclr/YuLEF20}
Wei Yu, Yichao Lu, Steve Easterbrook, et~al.,
\newblock ``Efficient and information-preserving future frame prediction and
  beyond,''
\newblock in {\em ICLR 2020}, 2020.

\bibitem{2021pdedriven}
J{\'e}r{\'e}mie Don{\`a}, Jean-Yves Franceschi, sylvain lamprier, et~al.,
\newblock ``Pde-driven spatiotemporal disentanglement,''
\newblock in {\em ICLR 2021}, 2021.

\end{thebibliography}

\end{document}